%% file: main.tex
\documentclass[conference]{IEEEtran}
\IEEEoverridecommandlockouts
\usepackage{cite}
\usepackage{graphicx}
\usepackage{xcolor}
\usepackage{amsmath}
\usepackage{wrapfig}
\usepackage{threeparttable}
\usepackage{float}
\usepackage{array}
\usepackage{multirow}
\usepackage{tabularx}
\usepackage{multicol}
\usepackage{etoolbox}
\usepackage{relsize}
\usepackage{float}
\usepackage{bbding}
\usepackage{pifont}
\usepackage{wasysym}
\usepackage{amssymb}
\usepackage{amsmath,amssymb,amsfonts}

\usepackage{algorithm}
\usepackage{algpseudocode}

\usepackage{graphicx}
\usepackage{textcomp}
\usepackage{xcolor}
\def\BibTeX{{\rm B\kern-.05em{\sc i\kern-.025em b}\kern-.08em
    T\kern-.1667em\lower.7ex\hbox{E}\kern-.125emX}}
\begin{document}


\title{Deployment of Large Language Models to Control Mobile Robots at the Edge\\

\thanks{\bf This work has been partially supported by NSF Award~\#~2201536.}
}


\author{Pascal Sikorski\textsuperscript{1}, Leendert Schrader\textsuperscript{2}, Kaleb Yu\textsuperscript{1}, Lucy Billadeau\textsuperscript{2}, Jinka Meenakshi\textsuperscript{3}, Naveena Mutharasan\textsuperscript{2}, \\ Flavio Esposito\textsuperscript{4}, Hadi AliAkbarpour\textsuperscript{5}, Madi Babaiasl\textsuperscript{6*}\thanks{\textsuperscript{1}Undergraduate Research Assistant, Computer Science Department, \textsuperscript{2}Undergraduate Research Assistant, Aerospace \& Mechanical Engineering Department, \textsuperscript{3}Research Assistant, Computer Science Department, \textsuperscript{4}Associate Professor, Computer Science Department, \textsuperscript{5}Assistant Professor, Computer Science Department, \textsuperscript{6*}Corresponding Author, Email: madi.babaiasl@slu.edu, CBL Assistant Professor of Robotics, Aerospace \& Mechanical Engineering Department, Saint Louis University, Saint Louis, USA}}

\maketitle

\begin{abstract}
This paper investigates the possibility of intuitive human-robot interaction through the application of Natural Language Processing (NLP) and Large Language Models (LLMs) in mobile robotics. This work aims to explore the feasibility of using these technologies for edge-based deployment, where traditional cloud dependencies are eliminated. The study specifically contrasts the performance of GPT-4-Turbo, which requires cloud connectivity, with an offline-capable, quantized version of LLaMA 2 (LLaMA 2-7B.Q5 K M). These results show that GPT-4-Turbo delivers superior performance in interpreting and executing complex commands accurately, whereas LLaMA 2 exhibits significant limitations in consistency and reliability of command execution. Communication between the control computer and the mobile robot is established via a Raspberry Pi Pico W, which wirelessly receives commands from the computer without internet dependency and transmits them through a wired connection to the robot's Arduino controller. This study highlights the potential and challenges of implementing LLMs and NLP at the edge, providing groundwork for future research into fully autonomous and network-independent robotic systems. For video demonstrations and source code, please refer to: https://tinyurl.com/MobileRobotGPT4LLaMA2024. 
\end{abstract}


\input{1_Introduction}
\input{2_Methods}

\begin{figure}
\includegraphics[scale=0.065]{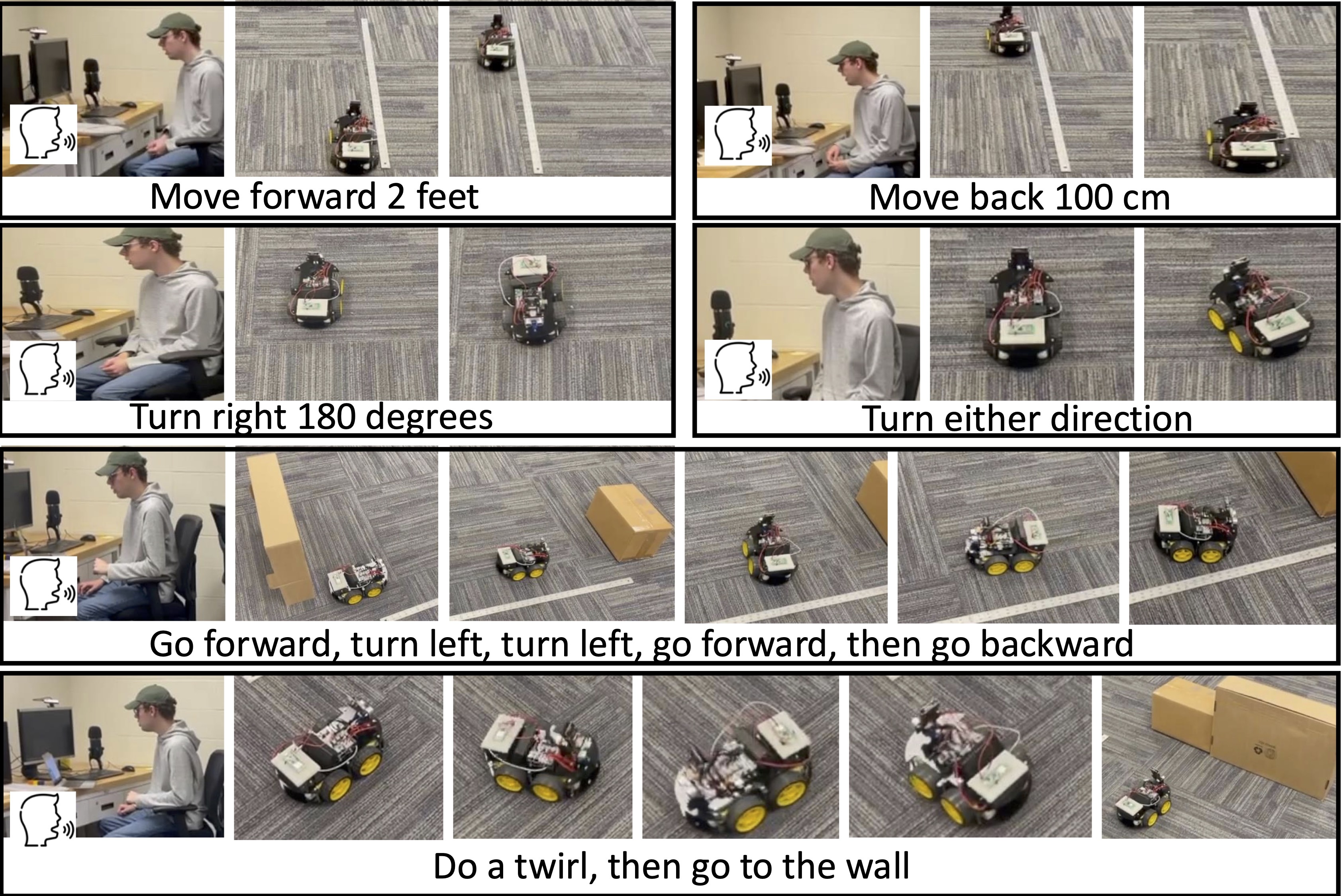}
 \centering
\caption{Some illustrative representations of the autonomous robotic response to diverse natural language commands. The sequence demonstrates the mobile robot's capability to interpret and execute commands ranging from basic directional instructions to complex navigational sequences. Each panel captures a moment in the process flow where the robot performs actions such as moving specific distances, executing precise turns, responding to ambiguous commands, and combining multiple maneuvers.} \label{fig:speech_to_robot_execution.jpg}
\end{figure}

\input{3_Results}

\input{4_Conclusion}

\bibliographystyle{IEEEtran} 
\bibliography{bibliography} 

\end{document}

%% file: 1_Introduction.tex
\section{Introduction}

\begin{figure*}
  \centering
  \includegraphics[scale=0.25]{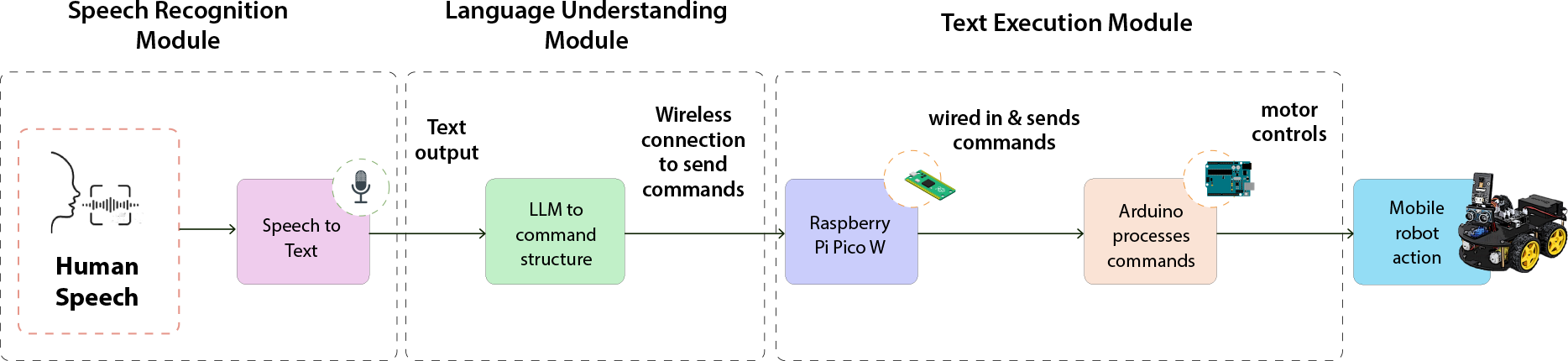}
  \caption{The schematic of the integrated mobile robot system architecture showing the workflow from voice input to robot action, including offline speech recognition, language understanding modules, and task execution via Raspberry Pi Pico W and Arduino.}
  \label{fig:architecture}
\end{figure*}

\subsection{Context and Background}

In recent years, the integration of robotics and artificial intelligence (AI) has started a new era in automation and intelligent systems \cite{soori2023artificial}. Mobile robotics, influenced by this integration, has evolved from simple navigational tasks to complex interaction scenarios involving dynamic decision-making capabilities \cite{yepez2023mobile}. On the other hand, significant strides in natural language processing (NLP) and large language models (LLMs) have transformed how machines understand and respond to human language \cite{zhang2023large}.

NLP technologies have traditionally relied on cloud-based infrastructures due to their substantial computational requirements \cite{dong2024creating}. However, there is a growing shift towards deploying these technologies on edge computing platforms \cite{dong2024creating}. This transition addresses growing concerns about privacy \cite{tallat2023navigating} and the limitations posed by intermittent or unreliable internet connectivity, critical in rescue operations or assistive technologies \cite{poirier2019voice}. By processing data locally, edge computing significantly reduces latency and enhances data security \cite{ahmadvand2021big} that meets the urgent needs of real-time robotic processing.

Integrating Language models and NLP in robotics can facilitate the robot's ability to interact fluidly with humans, learn from verbal instructions, and execute complex tasks \cite{zhang2023large}. This involves not only understanding spoken commands but also contextualizing actions and outcomes within a physical environment. Furthermore, the use of language models in an offline edge environment addresses significant concerns regarding data security and operational reliability in sensitive or unpredictable environments where constant internet connectivity is unfeasible \cite{ahmadvand2021big}. This approach also mitigates issues of latency, which are critical in scenarios where real-time robotics control is required, such as in navigation, obstacle avoidance, and interactive tasks.

The case study presented in this paper explores the practical deployment of LLMs and NLP in controlling a mobile robot and investigates the feasibility of using state-of-the-art LLMs on edge devices. This study aims to provide insights into the complexities and solutions associated with implementing NLP and LLMs in mobile robotics, paving the way for future research in these areas.

\subsection{Problem Statement}

Enhancing human-robot interaction through LLM and NLP poses substantial challenges, particularly in scenarios requiring robust privacy measures, such as assistive robotics and high-stake decision-making. Currently, most advanced robotic systems rely heavily on cloud computing for processing complex tasks \cite{vermesan2020internet}, which introduces significant vulnerabilities related to latency and data privacy. Furthermore, the dependency on continuous internet connectivity can hinder operational capabilities in disaster-stricken or remote areas. This study investigates the first step to a solution for intuitive human-robot interaction without the need for continuous cloud connectivity. 

\subsection{Research Objectives}

The first objective of this research is to study the feasibility of deploying LLMs and NLP for intuitive human-robot interaction within the context of mobile robots. The second objective is to investigate the potential and outline the limitations of utilizing such models in scenarios where cloud connectivity is compromised or entirely unavailable, moving controlling mobile robots towards self-sufficient systems without relying on remote servers. 

\subsection{Related Work}

The integration of LLMs and NLP into robotic systems is a new and exciting research area that aims to enhance robotic interaction capabilities in complex environments \cite{zhang2023large}. The rise of LLMs like GPT (Generative Pre-trained Transformer) \cite{radford2019language} further enhance NLP's potential in robotics \cite{wang2024large,zeng2023large}. These models, trained on vast amounts of text data, can understand and generate human-like text, allowing for more nuanced and complex interactions between humans and robots. Research groups have started using LLMs to control robots and autonomous vehicles \cite{sharan2023llm}. Liu et al. \cite{liu2023llm} proposed a human-robot collaboration framework based on LLM for manipulation tasks, while Sharan et al. \cite{sharan2023llm} utilized LLMs for self-driving vehicle planning.

Despite these promising research developments, a significant persisting challenge is the substantial computing resources required by LLMs, which can impede their integration into mobile robotic platforms. However, recent studies have shown that LLMs can be pruned by up to 40\% of their original size without noticeable performance loss \cite{gromov2024unreasonable}. This suggests a potential pathway for their integration into more compact systems and edge devices to be used in mobile robots. 

The current ongoing research aims to build on this research by exploring the potential for implementing LLMs and NLP into edge devices to control mobile robots. This paper marks the first step toward integrating NLP technologies with autonomous robotic functionalities on edge platforms. Although popular language models like GPT provide online API access for LLM integration, they rely on continuous connectivity to OpenAI's servers. This raises concerns about network reliability and privacy. Offline LLMs can address these issues by processing data locally on an edge device. Additionally, to demonstrate the feasibility of controlling mobile robots using LLMs and NLP, this paper shows how an edge device could be used to execute robotic commands. Leveraging the power of LLMs, natural human speech is converted into a format that is then processed locally on the mobile robot. 

%% file: 2_Methods.tex
\section{Methods}

\subsection{System Overview}

The core architecture of this system integrates a mobile robot with language processing capabilities, enabling it to perform tasks based on verbal commands. This work also investigates the feasibility of edge-based deployment of LLMs, which allows for real-time processing without the need for cloud connectivity (See Fig. \ref{fig:architecture}). This system includes the following components. An offline speech recognition module named VOSK by Alpha Cephei was used to convert spoken language into text. For the language understanding module, OpenAI's GPT-4-Turbo and a quantized version of Meta's LLaMa 2-7B called LLaMA 2-7B.Q5\_K\_M were utilized and compared. The differences between these models will be compared and contrasted based on their accuracy in speech interpretation. The key difference between the two models is that the LLaMa 2-7B model is running offline on a machine and does not require network access to generate responses. This shows a significant reason to use these models, as this will make the processing of the user's speech entirely local. These models process the transcribed text from the speech recognition software into appropriate tasks for the robot. This goes wirelessly to the task execution module, which translates the tasks into actionable commands that control the robot's hardware (motors and sensors) to interact with its environment (See also Algorithm alg:LLMalgo).

\begin{algorithm}
\caption{Robotic Control with LLM Model}
\label{alg:LLMalgo}
\begin{algorithmic}[1]
\State \textbf{Input:} User's Spoken Commands 
\State \textbf{Output:} Robot Executing Understandable Command
\newline

\Procedure{Robot + LLM Model}{}
    \State \textbf{Initialize system components:}
        \State \quad a. \text{Speech recognition module (VOSK)}
        \State \quad b. \text{NLP models (GPT-4-Turbo or LLaMA 2)}
        \State \quad c. \text{Mobile robot (Pico W and Arduino controller)}
        \State \quad d. \text{Form wireless connection between PC and robot}

    \State \State \textbf{Handle user input:}
        \State \quad a. \text{Begin listening for user input}
        \State \quad b. \text{Convert spoken command to text using VOSK}
        
    \State \State \textbf{Translate command through LLM model:}
        \If{GPT-4-Turbo}
            \State \text{Forward text command to OpenAI API}
            \State $\text{LLM}_{\text{Return}} \gets \text{Returned OpenAI API Value}$
        \ElsIf{LLaMA 2}
            \State \text{Forward text command to Local LLaMA 2}
            \State $\text{LLM}_{\text{Return}} \gets \text{Returned Local LLaMA 2 Value}$
        \EndIf
        \If{$\text{LLM}_{\text{Return}}$ is \textbf{Valid}}
                \State $\text{Robot}_{\text{Command}} \gets \text{LLM}_{\text{Return}}$
        \EndIf

    \State \State \textbf{Transmit the robot commands}
    \If{Wireless Connection is \textbf{Valid}}
        \State \text{Transmit string $\text{Robot}_{\text{Command}}$ across connection}

    \EndIf

    \State \State \textbf{Execute robot commands}

        \State $\text{List}_{\text{Command}} \gets \text{$\text{Robot}_{\text{Command}}$ separated by command}$
    
        \For{each command $i$ in $\text{List}_{\text{Command}}$}
            \State \text{Execute command $i$ as found in Arduino driver}
            \State \text{Wait for command $i$ to finish}
        \EndFor

\EndProcedure
\end{algorithmic}
\end{algorithm}

\subsection{Experimental Setup}

\begin{figure}
\includegraphics[scale=0.09]{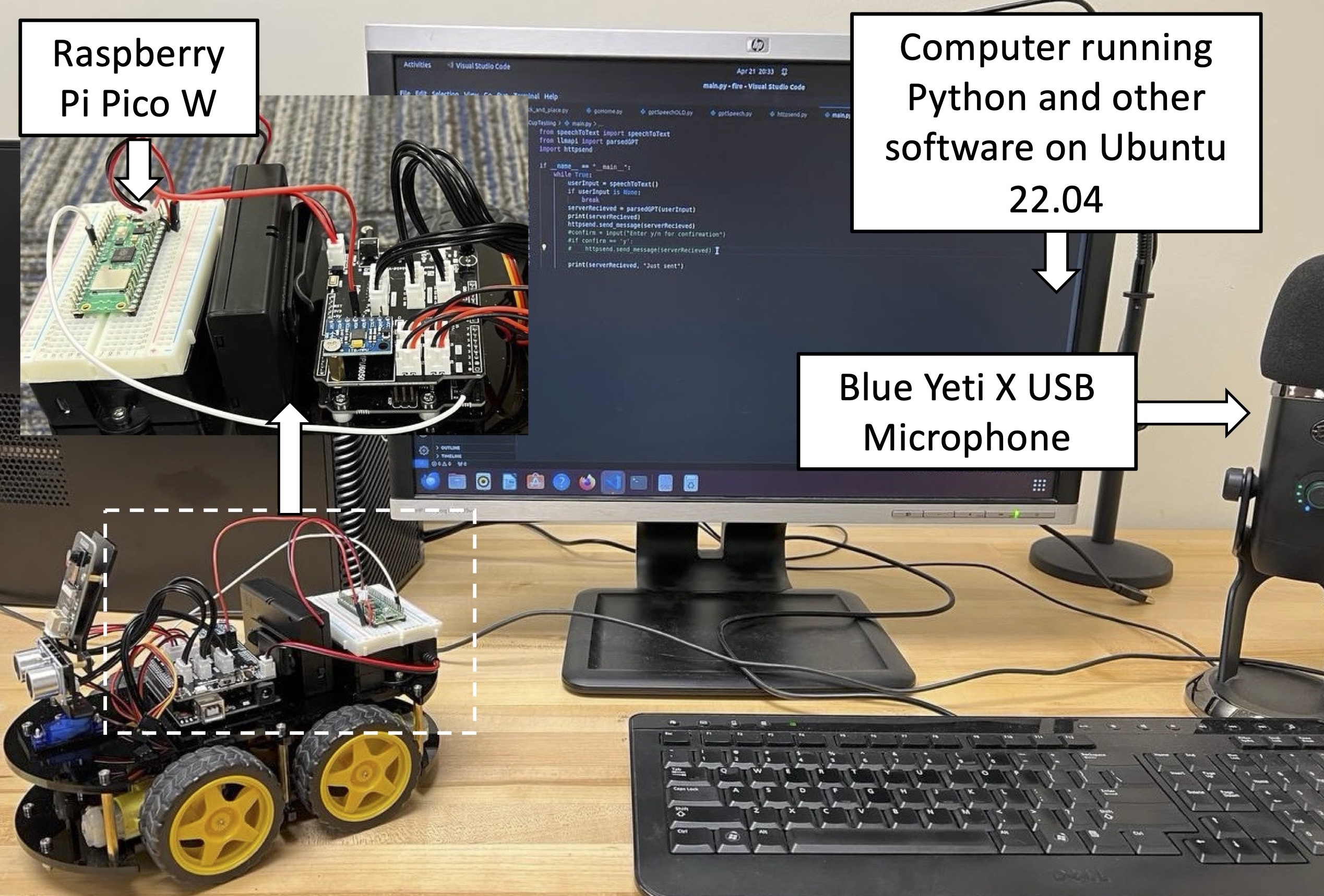}
 \centering
  \caption{Experimental setup. A Dell Precision 3660 Tower serves as the computing device running Python on Ubuntu 22.04 to process speech commands captured by a Blue Yeti X USB Microphone. The processed commands are wirelessly sent to a Raspberry Pi Pico W, which relays them to the robot via a Serial connection. The robot is controlled through its modified Arduino Uno board. The whole setup demonstrates the feasibility of edge-based command execution in mobile robotics.}
    \label{fig:experimental setup}
\end{figure}

Pertaining to hardware, a low-cost wheeled mobile robot named ELEGOO Smart Car V4.0 is utilized \cite{elegoo2024smart}. The reason this particular robot was chosen is its balance of affordability and the set of features that are conducive to work. Though, in attempting to establish WiFi connectivity with the Smart Car's proprietary Arduino Uno plus I/O Expansion Board setup, it was found that direct Wifi capabilities were highly restricted by the provided hardware, and it was best to outsource this connectivity into a Raspberry Pi Pico W microcontroller. Therefore, the flow of information was as follows: through WiFi, the user controller (desktop computer running Python on Ubuntu 22.04) sent commands to the Raspberry Pi Pico W over a WiFi server established from the Pico W, then the Pico W transmitted said commands through a UART Serial connection of the board and the mobile robot's Arduino Uno Board. The Pico W was powered by an external 6V battery pack, with a common ground being established between the two microcontrollers by way of an available grounding pin on the robot's I/O Expansion Board. The computer used is a Dell Precision 3660 Tower Core i7 with 32GB DDR5 up to 4400MHz UDIMM non-ECC memory and Nvidia RTXA4000 GPU. For audio capture, a Blue Yeti X USB Microphone is used to ensure high-quality input for the speech-to-text conversion process. Fig. \ref{fig:experimental setup} shows the experimental setup. 


\subsection{Mobile Robot Calibration}

\subsubsection{Motion Calibration of the Wheeled Mobile Robot ‐ Forward, Backward, Turning, and Stopping} 
\begin{figure}
\includegraphics[scale=0.13]{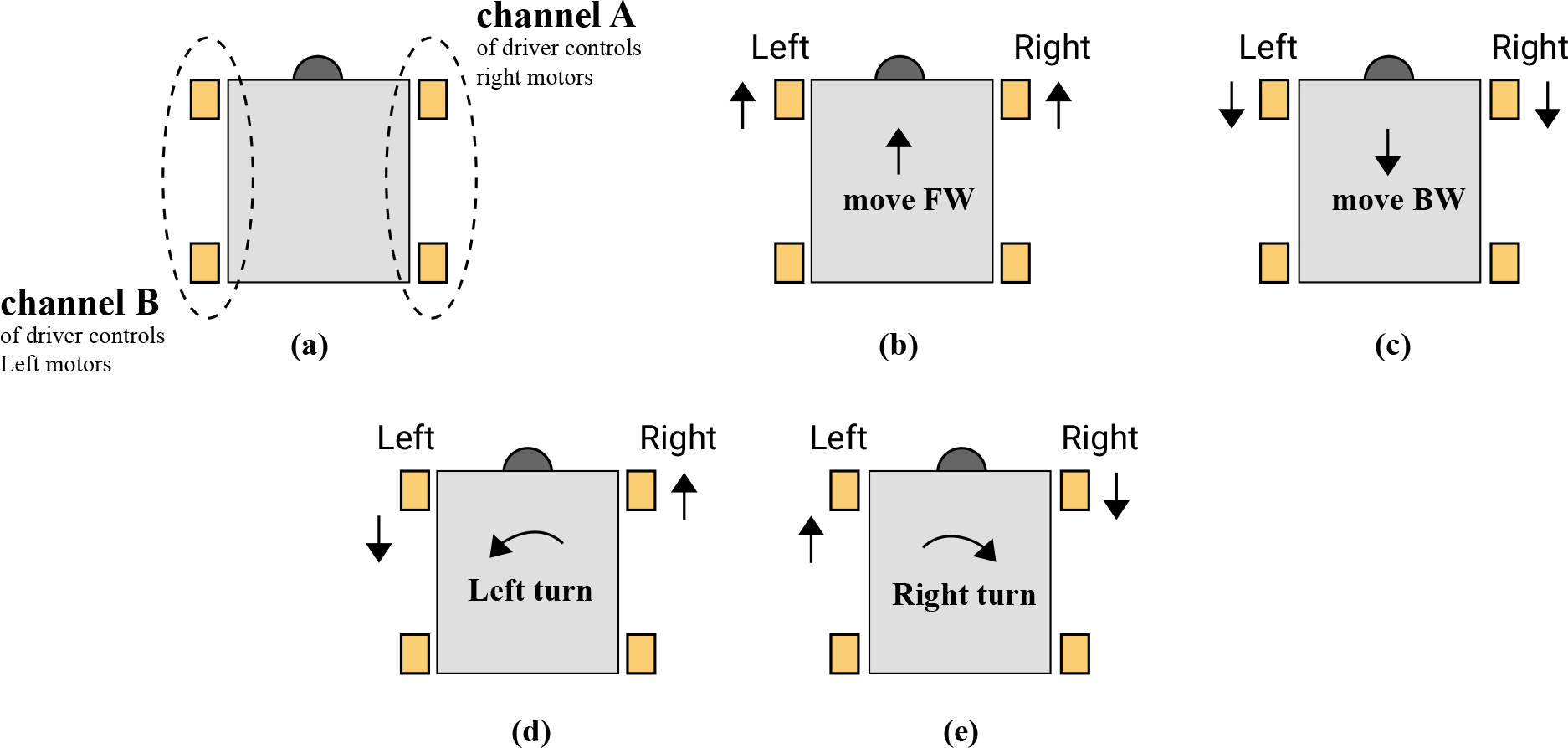}
 \centering
    \caption{Diagram illustrating the fundamental motions for the wheeled mobile robot with differential steering: Motor Driver Channels (a), Moving Forward (b), Moving Backward (c), Turning Left (d), and Turning Right (e).}
    \label{fig:robot_movements}
\end{figure}

Effective motion calibration is crucial for the autonomous operation of the robot. This calibration process addresses the four fundamental movements: forward, backward, turning (left and right), and stopping. The wheeled robot presented has 4 wheels. The right wheels are paired to operate together, as are the left wheels. The robot is equipped with a TB6612FNG motor driver that can control two DC motors. Therefore, one channel of the motor driver controls the right wheels and the other channel controls the left wheels (Fig. \ref{fig:robot_movements} - a). 

To move forward, both pairs of wheels (left and right) are driven forward, meaning both motors are activated to move in a direction that pushes the car forward (Fig. \ref{fig:robot_movements} - b). Conversely, to move backward, the motors are activated in the opposite direction, pulling the car backward (Fig. \ref{fig:robot_movements} - c). This is achieved by setting the motor driver input pins (IN1 and IN2 for one set of wheels, IN3 and IN4 for the other set) to the appropriate logic levels as specified in Table \ref{tab:combined-motor-config}. 

For executing turns with a mobile robot, the basic principle involves creating a differential in wheel speeds or directions between the left and right sides of the vehicle. This differential steering mechanism allows the robot to navigate left or right turns. Here, left and right turns for the robot are implemented by changing the direction in which each set of wheels moves. For the robot to turn left, the wheels are controlled such that the left wheels move backward and the right wheels move forward (Fig. \ref{fig:robot_movements} - d). For the robot to turn right, the opposite action is taken, meaning that the left wheels move forward and the right wheels move backward (Fig. \ref{fig:robot_movements} - e). The logic levels for the motor driver input pins are specified in Table \ref{tab:combined-motor-config}.  


\begin{table}[h]
\centering
\caption{Motor Driver Logic for Mobile Robot Movement Control}
\label{tab:combined-motor-config}
\renewcommand{\arraystretch}{1.1}
\setlength{\tabcolsep}{7pt}
\tiny
\begin{tabular}{|c|c|c|c|c|c|c|}
\hline
Right Wheels & Left Wheels & Robot Movement & IN1 & IN2 & IN3 & IN4 \\ \hline
FW & FW & Forward & 1 & 0 & 1 & 0 \\
BW & BW & Backward & 0 & 1 & 0 & 1 \\
BW & FW & Right Turn & 0 & 1 & 1 & 0 \\
FW & BW & Left Turn & 1 & 0 & 0 & 1 \\ \hline
\end{tabular}
\end{table}

To stop the robot, both the right wheels and left wheels should be stopped, and this can be done by setting the logic level of the PWM (Pulse Width Modulation) pins of the motor driver to 0. 

For linear Motion Calibration (Forward and Backward), the robot is programmed to traverse a fixed track at different PWM outputs to capture the baseline performance data. Distances traveled over incremental time intervals were recorded. A trend line was then fitted to the distance-time data for each PWM value, and the corresponding linear speed was calculated from the slope of the line. The best fit for the PWM-speed data is then determined to establish the relationship between speed and the corresponding PWM value that should be written to the PWM pins to achieve the desired speed. This relationship was modeled by fitting a second-degree polynomial to the speed and PWM data, resulting in an equation that predicts the necessary PWM value for a given speed. The polynomial fit for the PWM as a function of speed for the forward movement is given by: $\text{PWM} = -0.0264 (\text{Speed})^2 + 5.4266(\text{Speed}) - 35.889$.


A similar equation can be found for the backward movement. The time that the motors should stay on can be calculated as follows: $\text{Time} = \frac{\text{Distance}}{\text{Speed}}$.




Following the methodology used for linear motion calibration, a similar process is applied to angular motion calibration (turning maneuvers). The goal was to characterize the relationship between PWM outputs and angular speeds (degrees per second) as the robot rotates in place either to the right or left. For data collection, the robot was programmed to rotate in place at different PWM outputs. The angles turned were recorded over specified time intervals. After that, for each PWM setting, the angle turned was plotted against time, and the angular speed was calculated from the slope of the trend line for each time interval. Finally, a second-order polynomial was fitted to model the angular speed as a function of PWM. This model helps in predicting the PWM value required to achieve a desired angular speed. 



The equation predicting the amount of PWM needed for motors based on the angular speed (deg/sec) is $\text{PWM} = 0.001 (\text{AngularSpeed})^2 + 0.095(\text{AngularSpeed}) + 92.3$.


A similar equation can be found for the left turn, and then the time that the motors should stay on for the precise turn can be calculated as follows: $\text{Time} = \frac{\text{Angle}}{\text{Angular Speed}}$.


\subsubsection{Calibration and Denoising the Ultrasonic Sensor}

\begin{figure}
\includegraphics[scale=0.065]{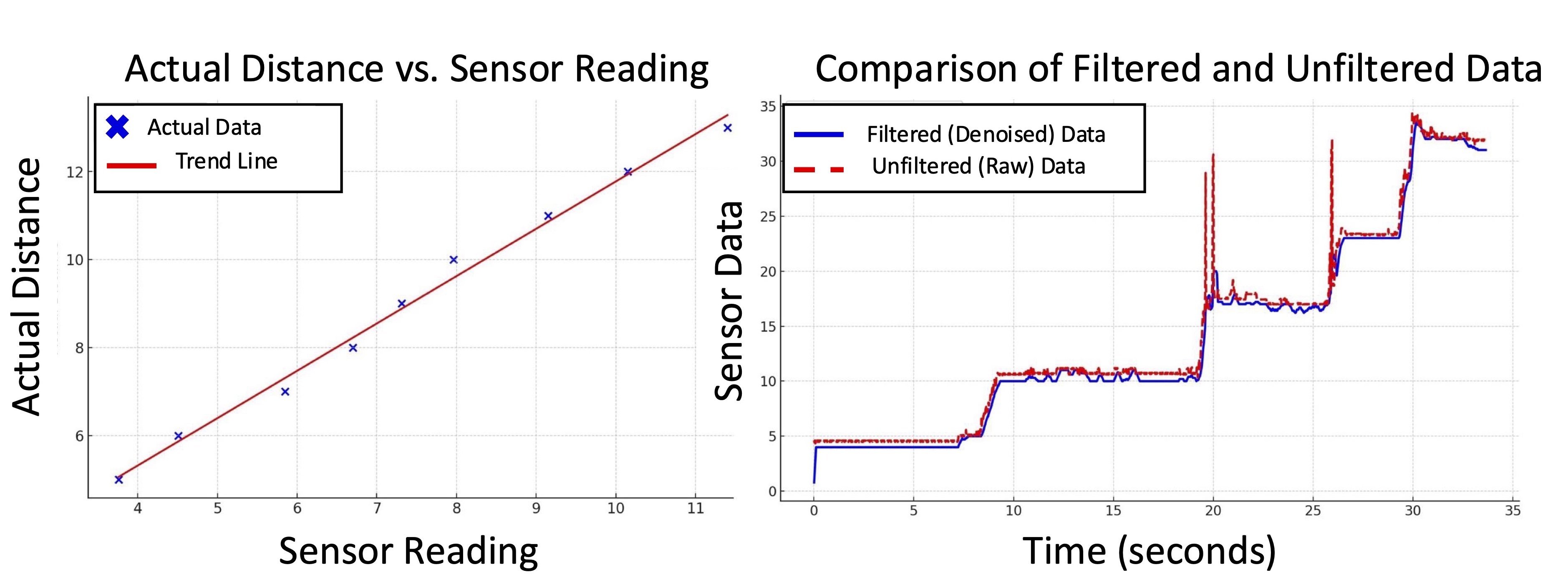}
 \centering
    \caption{Calibration and Noise Filtering of Ultrasonic Sensor Data. The left graph illustrates the linear relationship between the actual distance and the ultrasonic sensor readings with a trend line, while the right graph displays the effectiveness of a noise-filtering process on ultrasonic sensor data over time. Note that, the distance of the object in front of the sensor gradually increased over time.}
    \label{fig:ultrasonic_sensor_calibration_denoising}
\end{figure}

For obstacle detection and avoidance, an HC-SR04 ultrasonic sensor was used, which measures distances by emitting ultrasonic waves and calculating the time taken for the echo to return. This section details the calibration and noise removal processes implemented to enhance the sensor's accuracy and reliability.

The primary objective of sensor calibration was to correlate the sensor's raw measurements with actual distance measurements. A series of experiments were performed where the sensor's distance readings were recorded against known distances, and measured manually. These data points were then used to establish a linear regression model that predicts the actual distance ($y$) as a function of the sensor reading ($x$). The calibration model is defined by the linear equation $\hat{y} = mx + b$, where $m$ is the slope and $b$ is the y-intercept. These parameters were calculated using the least squares method, aiming to minimize the sum of squared differences between the observed distances and those predicted by the model. Fig. \ref{fig:ultrasonic_sensor_calibration_denoising} - Left shows the actual distance versus the sensor reading along with the trend line. The relationship between the actual distance and the sensor reading can be expressed by the following equation: $\text{Actual Distance} = 1.0759 \times \text{Sensor Reading} + 1.0158$.


To address the noise inherent in the raw ultrasonic sensor data, which can be caused by various factors including environmental conditions and sensor anomalies, a Simple Moving Average (SMA) filter is implemented. The SMA filter improves the reliability of distance measurements by averaging several consecutive readings, thereby smoothing out random fluctuations. Mathematically, SMA can be expressed as: ${SMA}_i = \frac{1}{M} \sum\limits_{j=0}^{M-1}d_{i+j}$.


Where, ${\text{SMA}}_i$ is the simple moving average at the $i^{th}$ position in the smoothed series, M is the total number of data points in the window used to calculate the average, including the current point $d_i$ and the next M-1 points, and $d_{i+j}$ represents each data point in the window, starting from the current data point $d_i$ and including the next M-1 points in the series. A window size of 5 is chosen for the SMA, as this size provides a good balance between noise reduction and response speed. The filter operates by maintaining a rolling window of the last five calibrated distance readings and computing their average. This smoothed value is then used as the output distance reading, reducing the impact of transient noise spikes. Fig. \ref{fig:ultrasonic_sensor_calibration_denoising} - Right shows the smoothed distance readings obtained using the 5-point SMA filter that clearly shows the reduction in noise spikes and the preservation of the overall trend in the data.

\subsection{Raspberry Pi Pico W Access Point and Arduino}

A Raspberry Pi Pico W is set up to serve as an access point over Wifi. The socket receives a string which is sent to the Arduino using microPython UART Serial communication initialization. The specific Pico W-to-Arduino communication was carried out through a wired Serial protocol at 9600 baud rate using the hardware Serial receiver (RX) of the Arduino tethered to the microPython-initialized UART transmission line (TX) of the Pico W. A common ground was established between the Pico W and Arduino via connection from the servo GND pin on the Arduino I/O Expansion Board to the GND pin of the Pico W associated with the power supply. Once hardware connectivity was established, the Pico W was programmed in microPython to send all received commands over Wifi directly over UART to the Arduino; all mobility interpretation was coded directly in the Arduino, with the Pico W acting as a receiver-transmitter without any internal command transformations. This line of communication ensures a network between the client and the mobile robot.

\subsection{LLM Integration}

This project involves a language model as the central point. The OpenAI API was utilized to access GPT-4-Turbo. The overall responses are accurate and quick to generate. However, this model does require network access and a payment method. LLaMa 2 7B is utilized through a program called "llama.cpp" which allows users to easily download and utilize different models on their machine. The hardware used was incompatible with official base models due to GPU VRAM limitations, leading to the decision to utilize a quantized version of Llama 2 called LLaMA 2-7B.Q5\_K\_M. This quantized model enables the language model to run offline and on the available hardware. The models were given instructions to convert a given string of text into a command followed by distance and/or sequences of commands. The model was instructed to interpret the naturally spoken language and deduce the closest set of commands that would match the intentions of the initial input. For example, the words "Go forward 100cm" would be spoken into a microphone and sent to one of the chosen Language models. The model would output the command format which if accurate, would result in "f,100". These commands would then be wirelessly transmitted to the Raspberry Pi Pico W. The output of commands generated from these models was evaluated to determine their accuracy in controlling the smart car through the resulting actions.

%% file: 3_Results.tex
\section{Results}

From user command to calibrated execution, a plethora of actions were able to be consistently executed in line with the expectations of natural language input. Precise directional, translational, and angular mobility was accurately interpreted, communicated, and employed at each stage of the information flow. The mobile robot was also uniquely prepared for indirect commands, being able to follow complex sequences transmitted by the user's speech. Obstacle avoidance, indefinite motor activation, precision directional control, travel sequencing, as well as complex combinations of these functions were seamlessly integrated into the mobility of the robot from simple, intuitive user speech (see Fig. \ref{fig:speech_to_robot_execution.jpg}).

Integrating GPT-4-Turbo with the mobile robot has shown extremely strong results. The mobile robot can understand a mix of simple and complex instructions to properly execute desired commands in most contexts. GPT-4-Turbo has shown its ability to critically translate provided prompts to fit the parameters of robot functionality, such as to "move forward 2 feet". In this example, GPT-4-Turbo needs to understand and convert 2 feet into centimeters for proper function, as it properly does with forward output between 60 and 61 centimeters. When increasing the complexity of prompts, GPT-4-Turbo still has impressive performance. In a more complex prompt such as "Do a twirl, then go to the wall", success is still achieved. GPT-4-Turbo will still correctly translate into correct functionality, rotating 360 degrees, then moving forward (understanding to utilize its stopping mechanism from the onboard ultrasonic sensor) to stop at a wall.

However, in testing with the quantized model of LLaMa 2 7B, the offline results were found to be subpar. Results taken from prompts as basic as to "Move forward 50 centimeters" often resulted in cluttered outputs, such as moving forward 450 centimeters in added sequence, or even moving forward only 10 centimeters. Similar results can be found across all trials using the LLaMa 2 7B model. These results show that the implementation of a more accessible, lower-computational cost, offline model still struggles to produce reliable output for robotic control in real-world situations.

Table \ref{tab:command-comparison} demonstrates the performance comparison between GPT-4-Turbo and quantized version of LLaMA 2 across a series of command execution trials.


\begin{table}[H]
\centering
\caption{Command Execution: GPT-4-Turbo vs LLaMA 2 7B.Q5\_K\_M Quantized}
\label{tab:command-comparison}
\renewcommand{\arraystretch}{1.1}
\setlength{\tabcolsep}{9.6pt}
\tiny
\begin{tabular}{|c|c|c|c|c|c|c|}
\hline
\multicolumn{1}{|c|}{\textbf{Command}} & \multicolumn{3}{c|}{\textbf{GPT-4-Turbo}} & \multicolumn{3}{c|}{\textbf{LLaMA 2 Quantized}} \\ \hline
\textbf{No.*} & \textbf{Trial 1} & \textbf{Trial 2} & \textbf{Trial 3} & \textbf{Trial 1} & \textbf{Trial 2} & \textbf{Trial 3} \\ \hline
1 & PASS & PASS & PASS & \textcolor{red}{FAIL} & \textcolor{red}{FAIL} & \textcolor{red}{FAIL} \\
2 & PASS & PASS & PASS & \textcolor{red}{FAIL} & \textcolor{red}{FAIL} & \textcolor{red}{FAIL} \\
3 & PASS & PASS & PASS & \textcolor{red}{FAIL} & PASS & \textcolor{red}{FAIL} \\
4 & PASS & PASS & PASS & \textcolor{red}{FAIL} & PASS & \textcolor{red}{FAIL} \\
5 & PASS & PASS & PASS & PASS & \textcolor{red}{FAIL} & \textcolor{red}{FAIL} \\
6 & PASS & PASS & PASS & \textcolor{red}{FAIL} & PASS & \textcolor{red}{FAIL} \\
7 & PASS & PASS & PASS & PASS & \textcolor{red}{FAIL} & \textcolor{red}{FAIL} \\
8 & PASS & PASS & PASS & \textcolor{red}{FAIL} & \textcolor{red}{FAIL} & \textcolor{red}{FAIL} \\
9 & PASS & PASS & PASS & \textcolor{red}{FAIL} & \textcolor{red}{FAIL} & \textcolor{red}{FAIL} \\
10 & \textcolor{red}{FAIL} & \textcolor{red}{FAIL} & \textcolor{red}{FAIL} & \textcolor{red}{FAIL} & \textcolor{red}{FAIL} & \textcolor{red}{FAIL} \\
11 & PASS & \textcolor{red}{FAIL} & PASS & \textcolor{red}{FAIL} & \textcolor{red}{FAIL} & \textcolor{red}{FAIL} \\
12 & PASS & PASS & PASS & \textcolor{red}{FAIL} & \textcolor{red}{FAIL} & \textcolor{red}{FAIL} \\
13 & PASS & PASS & PASS & \textcolor{red}{FAIL} & \textcolor{red}{FAIL} & \textcolor{red}{FAIL} \\
14 & PASS & PASS & PASS & PASS & \textcolor{red}{FAIL} & PASS \\
15 & PASS & PASS & PASS & \textcolor{red}{FAIL} & \textcolor{red}{FAIL} & \textcolor{red}{FAIL} \\
16 & PASS & PASS & PASS & \textcolor{red}{FAIL} & \textcolor{red}{FAIL} & \textcolor{red}{FAIL} \\
17 & PASS & PASS & PASS & \textcolor{red}{FAIL} & PASS & \textcolor{red}{FAIL} \\
18 & PASS & PASS & PASS & \textcolor{red}{FAIL} & \textcolor{red}{FAIL} & \textcolor{red}{FAIL} \\
19 & PASS & PASS & PASS & \textcolor{red}{FAIL} & \textcolor{red}{FAIL} & PASS \\
20 & PASS & PASS & PASS & \textcolor{red}{FAIL} & \textcolor{red}{FAIL} & \textcolor{red}{FAIL} \\
21 & PASS & PASS & \textcolor{red}{FAIL} & \textcolor{red}{FAIL} & \textcolor{red}{FAIL} & \textcolor{red}{FAIL} \\
22 & \textcolor{red}{FAIL} & \textcolor{red}{FAIL} & \textcolor{red}{FAIL} & \textcolor{red}{FAIL} & \textcolor{red}{FAIL} & \textcolor{red}{FAIL} \\
23 & \textcolor{red}{FAIL} & PASS & \textcolor{red}{FAIL} & \textcolor{red}{FAIL} & \textcolor{red}{FAIL} & \textcolor{red}{FAIL} \\ \hline
\end{tabular}
\begin{tablenotes}[flushleft]
\item * Command 1: Move forward 300 cm, Command 2: Move forward 2 feet, Command 3: Move back 100 cm, Command 4: Move forward, Command 5: Move forward 50 cm, Command 6: Move backward 25 cm, Command 7: Turn right, Command 8: Turn right 180 degrees, Command 9: Turn left pi radians, Command 10: Turn, Command 11: Turn either direction, Command 12: Turn around, Command 13: Twirl, Command 14: Move forward 100 then stop moving, Command 15: Go forward then go backwards, Command 16: Move forward then come back, Command 17: Go forward, turn left, turn left, go forward, then go backward, Command 18: Do a twirl, then go to the wall, Command 19: Utilize your ultrasonic sensor, Command 20: Go behind you, then come back to where you started, Command 21: Move to the left, go to the wall, then come back, Command 22: Pick a route to traverse around the room, Command 23: Traverse around the room with lots of moves. 
\end{tablenotes}
\end{table}

%% file: 4_Conclusion.tex
\section{Conclusion and Future Work}

This paper presented a comparative analysis of the efficacy of GPT-4-Turbo and the quantized LLaMA 2 models for integrating NLP and LLMs into mobile robotics. The research highlighted GPT-4-Turbo's proficiency in handling complex commands effectively when cloud connectivity is available. In contrast, the quantized LLaMA 2 model, despite being operable offline, exhibited limitations in reliability and output consistency. In the comparison of GPT-4-Turbo and the quantized LLaMA 2 models, a passing accuracy of 85\% was observed in the GPT-4-Turbo model, with only 13\% observed in the LLaMA 2 model, based on test instructions. This accuracy with instructions can be derived from models tested as shown in Table \ref{tab:command-comparison}.

The experimental framework, utilizing a Raspberry Pi Pico W for wireless communication from a control computer, and wired communication to an Arduino-driven robot, exemplifies an approach to edge-based robotic control. The exploration of offline, edge-based NLP for mobile robots provides a foundation for future research aimed at refining offline models for real-world robotic instruction and applications. Successful integration of a quantized model, even with it's limitations, presents a new method of optimizing lower-computational cost models to operate independent from cloud connectivity. This work contributes to fields of autonomous robotics, demonstrating practical methods using NLP to solve robotic control problems in environments where privacy and real-time performance are crucial.

Looking forward, future work will be dedicated to specifically fine-tuning offline language models for robotic applications. The goal is to enhance the model's reliability and consistency by tailoring it closely to the specific needs and constraints of robotic interaction. This process will include pruning and optimizing the model to ensure it operates efficiently on edge devices. The underperformance of the LLaMA 2 model observed in the experiments is largely attributed to its generic configuration, which was not initially optimized for specific robotic tasks. By adjusting the model parameters and structure to better fit the practical requirements of mobile robots, significant performance improvements are anticipated. These modifications will enable the deployment of sophisticated NLP capabilities directly on robots, facilitating more dynamic and context-aware interactions between humans and robots without relying on cloud connectivity. Crucially, this development will pave the way towards a fully offline operational model, eliminating dependencies on cloud connectivity.

\vspace{-1mm}
\section*{Acknowledgement}
\vspace{-2mm}
This work has been partially supported by NSF Award~\#~2201536. 

%% file: main.bbl
\begin{thebibliography}{10}
\providecommand{\url}[1]{#1}
\csname url@rmstyle\endcsname
\providecommand{\newblock}{\relax}
\providecommand{\bibinfo}[2]{#2}
\providecommand\BIBentrySTDinterwordspacing{\spaceskip=0pt\relax}
\providecommand\BIBentryALTinterwordstretchfactor{4}
\providecommand\BIBentryALTinterwordspacing{\spaceskip=\fontdimen2\font plus
\BIBentryALTinterwordstretchfactor\fontdimen3\font minus
  \fontdimen4\font\relax}
\providecommand\BIBforeignlanguage[2]{{%
\expandafter\ifx\csname l@#1\endcsname\relax
\typeout{** WARNING: IEEEtran.bst: No hyphenation pattern has been}%
\typeout{** loaded for the language `#1'. Using the pattern for}%
\typeout{** the default language instead.}%
\else
\language=\csname l@#1\endcsname
\fi
#2}}

\bibitem{soori2023artificial}
M.~Soori, B.~Arezoo, and R.~Dastres, ``Artificial intelligence, machine
  learning and deep learning in advanced robotics, a review,'' \emph{Cognitive
  Robotics}, 2023.

\bibitem{yepez2023mobile}
D.~F. Y{\'e}pez-Ponce, J.~V. Salcedo, P.~D. Rosero-Montalvo, and J.~Sanchis,
  ``Mobile robotics in smart farming: current trends and applications,''
  \emph{Frontiers in Artificial Intelligence}, vol.~6, p. 1213330, 2023.

\bibitem{zhang2023large}
C.~Zhang, J.~Chen, J.~Li, Y.~Peng, and Z.~Mao, ``Large language models for
  human-robot interaction: A review,'' \emph{Biomimetic Intelligence and
  Robotics}, p. 100131, 2023.

\bibitem{dong2024creating}
Q.~Dong, X.~Chen, and M.~Satyanarayanan, ``Creating edge ai from cloud-based
  llms,'' in \emph{Proceedings of the 25th International Workshop on Mobile
  Computing Systems and Applications}, 2024, pp. 8--13.

\bibitem{tallat2023navigating}
R.~Tallat, A.~Hawbani, X.~Wang, A.~Al-Dubai, L.~Zhao, Z.~Liu, G.~Min, A.~Y.
  Zomaya, and S.~H. Alsamhi, ``Navigating industry 5.0: A survey of key
  enabling technologies, trends, challenges, and opportunities,'' \emph{IEEE
  Communications Surveys \& Tutorials}, 2023.

\bibitem{poirier2019voice}
S.~Poirier, F.~Routhier, and A.~Campeau-Lecours, ``Voice control interface
  prototype for assistive robots for people living with upper limb
  disabilities,'' in \emph{2019 IEEE 16th International Conference on
  Rehabilitation Robotics (ICORR)}.\hskip 1em plus 0.5em minus 0.4em\relax
  IEEE, 2019, pp. 46--52.

\bibitem{ahmadvand2021big}
H.~Ahmadvand, T.~Dargahi, F.~Foroutan, P.~Okorie, and F.~Esposito, ``Big data
  processing at the edge with data skew aware resource allocation,'' in
  \emph{2021 IEEE conference on network function virtualization and software
  defined networks (NFV-SDN)}.\hskip 1em plus 0.5em minus 0.4em\relax IEEE,
  2021, pp. 81--86.

\bibitem{vermesan2020internet}
O.~Vermesan, R.~Bahr, M.~Ottella, M.~Serrano, T.~Karlsen, T.~Wahlstr{\o}m,
  H.~E. Sand, M.~Ashwathnarayan, and M.~T. Gamba, ``Internet of robotic things
  intelligent connectivity and platforms,'' \emph{Frontiers in Robotics and
  AI}, vol.~7, p. 104, 2020.

\bibitem{radford2019language}
A.~Radford, J.~Wu, R.~Child, D.~Luan, D.~Amodei, I.~Sutskever, \emph{et~al.},
  ``Language models are unsupervised multitask learners,'' \emph{OpenAI blog},
  vol.~1, no.~8, p.~9, 2019.

\bibitem{wang2024large}
J.~Wang, Z.~Wu, Y.~Li, H.~Jiang, P.~Shu, E.~Shi, H.~Hu, C.~Ma, Y.~Liu, X.~Wang,
  \emph{et~al.}, ``Large language models for robotics: Opportunities,
  challenges, and perspectives,'' \emph{arXiv preprint arXiv:2401.04334}, 2024.

\bibitem{zeng2023large}
F.~Zeng, W.~Gan, Y.~Wang, N.~Liu, and P.~S. Yu, ``Large language models for
  robotics: A survey,'' \emph{arXiv preprint arXiv:2311.07226}, 2023.

\bibitem{sharan2023llm}
S.~Sharan, F.~Pittaluga, M.~Chandraker, \emph{et~al.}, ``Llm-assist: Enhancing
  closed-loop planning with language-based reasoning,'' \emph{arXiv preprint
  arXiv:2401.00125}, 2023.

\bibitem{liu2023llm}
H.~Liu, Y.~Zhu, K.~Kato, I.~Kondo, T.~Aoyama, and Y.~Hasegawa, ``Llm-based
  human-robot collaboration framework for manipulation tasks,'' \emph{arXiv
  preprint arXiv:2308.14972}, 2023.

\bibitem{gromov2024unreasonable}
A.~Gromov, K.~Tirumala, H.~Shapourian, P.~Glorioso, and D.~A. Roberts, ``The
  unreasonable ineffectiveness of the deeper layers,'' \emph{arXiv preprint
  arXiv:2403.17887}, 2024.

\bibitem{elegoo2024smart}
ELEGOO, ``Smart robot car kit v4.0 (with camera),''
  \url{https://www.elegoo.com/products/elegoo-smart-robot-car-kit-v-4-0}, 2024,
  accessed: 2024-06-19.

\end{thebibliography}
